# Depth-Aware Rover: A Study of Edge AI and Monocular Vision for Real-World Implementation

Lomash Relia
Department of Computer Engineering
Devang Patel Institute of Advanced
Technology and Research
Anand, India
lomashrelia@gmail.com

Jai G Singla
Space Applications Centre
Indian Space Research Organization
Ahmedabad, India
jaisingla@sac.isro.gov.in

Amitabh
Space Applications Centre
Indian Space Research Organization
Ahmedabad, India
amitabh@sac.isro.gov.in

Nitant Dube
Space Applications Centre
Indian Space Research Organization
Ahmedabad, India
nitant@sac.isro.gov.in

*Abstract*— This study analyses simulated and real-world implementations of depth-aware rover navigation, highlighting the transition from stereo vision to monocular depth estimation using edge AI. A Unity-based lunar terrain simulator with stereo cameras and OpenCV's StereoSGBM was used to generate disparity maps. A physical rover built on Raspberry Pi 4 employed UniDepthV2 for monocular metric depth estimation and YOLO12n for real-time object detection. While stereo vision yielded higher accuracy in simulation, the monocular approach proved more robust and cost-effective in real-world deployment, achieving 0.1 FPS for depth and 10 FPS for detection.

## I. Introduction

Autonomous rover navigation in challenging terrains requires robust depth perception capabilities for obstacle detection and path planning. Traditional planetary rover systems, such as NASA's Mars rovers [1], [2], rely heavily on stereo camera configurations for depth estimation. However, these systems face significant limitations in real-world deployment, including sensitivity to calibration disturbances, performance degradation in low-texture environments, and computational complexity that often necessitates slow, cautious traversal with extensive ground-based monitoring.

The emergence of deep learning-based monocular depth estimation presents a promising alternative, offering potential advantages in terms of hardware simplicity, calibration robustness, and cost-effectiveness. However, the gap between simulation performance and real-world implementation remains poorly understood, particularly regarding edge AI deployment constraints and practical system integration challenges.

This paper addresses this knowledge gap through a systematic comparative study involving both Unity-based simulation and physical rover prototype development. Our contribution is threefold: (1) identification of key discrepancies between simulated and real-world stereo vision performance, (2) evaluation of monocular depth estimation viability for edge AI rover applications, and (3) practical insights for bridging the simulation-to-reality gap in autonomous navigation systems.

## II. Related Work

Stereo vision has been the dominant approach for rover depth estimation, with Semi-Global Block Matching (SGBM) [3] providing dense disparity maps with good accuracy-efficiency balance. However, these methods require precise calibration and suffer from fundamental limitations in textureless regions and varying illumination conditions commonly encountered in outdoor environments.

Recent advances in deep learning have enabled accurate monocular depth estimation through both supervised and self-supervised approaches. Methods like MiDaS [4] provide relative depth estimation with good generalization, while DepthAnything [5] offers robust zero-shot performance across diverse domains. For robotics applications requiring absolute measurements, metric depth estimation approaches are crucial. DepthPro [6] achieves exceptional accuracy but demands significant computational resources, while UniDepthV2 [7] offers a balanced approach with metric depth estimation suitable for edge deployment through architectural optimizations.

The deployment of computer vision algorithms on resource-constrained hardware presents unique challenges regarding inference speed and power consumption. The YOLO family of detectors has demonstrated exceptional performance on edge hardware, with YOLO12 [8] incorporating attention mechanisms while maintaining real-time performance suitable for robotics applications.

The transfer of computer vision systems from simulation to real-world deployment remains a significant challenge. Despite extensive research in individual components, several gaps remain: most depth estimation studies focus on benchmark datasets rather than practical robotics deployment, few studies systematically compare simulated and real-world performance for complete navigation systems, and the practical challenges of deploying state-of-the-art vision models on resource-constrained robotics platforms are underexplored. This work addresses these gaps by providing a comprehensive comparison between simulated and real-world implementations, focusing on practical deployment challenges for autonomous rover navigation.

## III. Methodology

This study aims to evaluate the practical challenges of implementing depth-aware navigation systems on planetary rovers by comparing controlled simulation with real-world deployment. The methodology consists of three interconnected stages: simulation-based development using Unity, construction and deployment of a physical rover prototype, and systematic performance evaluation across diverse testing environments.

*A. Simulation Setup*

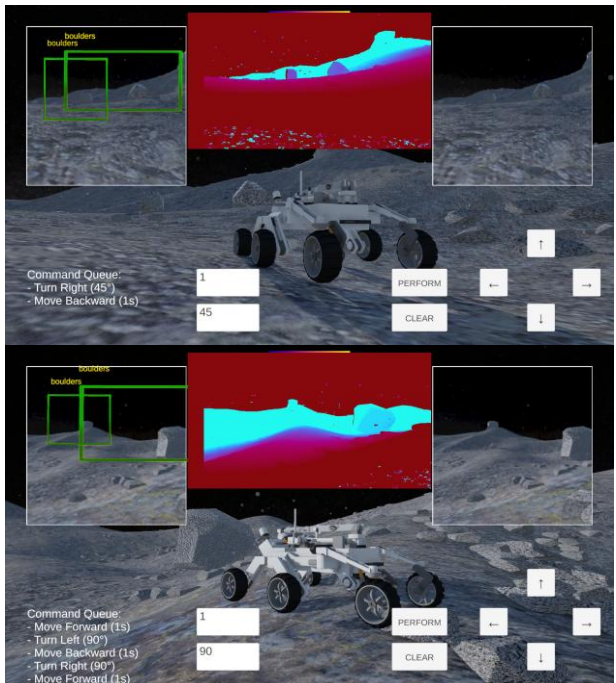

*Figure 1 Simulation on Stereovision based Rover Navigation*

A simulation environment was created in Unity 6 [9], utilizing the Lunar Landscape 3D asset to model a photorealistic lunar terrain [10]. The rover model was based on the Espacial Explorer T-30 concept design [11]. A stereo camera setup was mounted virtually on the rover, enabling the testing of visual perception pipelines. The virtual stereo camera system was configured with:

- Baseline: 24 Unity units (X-axis)
- Field of View: 60° × 60°
- Resolution: 500 × 500 pixels
- Focal Length: 433 pixels (derived from FoV and resolution)

Depth was computed using the formula, $Z = (f \times B)/d$, where f is the focal length, B is the baseline, and d is the disparity. Disparity maps were generated using OpenCV's StereoSGBM algorithm with default parameters, producing dense and reliable depth maps under optimal simulation conditions. To enable semantic perception, a custom-trained YOLO12n model was converted to ONNX format and deployed in Unity using com.unity.ai.inference [12]. The model was specifically trained to detect rover-relevant obstacles such as large boulders. When an obstacle appeared directly in the rover's path, the simulation halted execution, simulating the need for human oversight.

*B. Physical Rover Implementation*

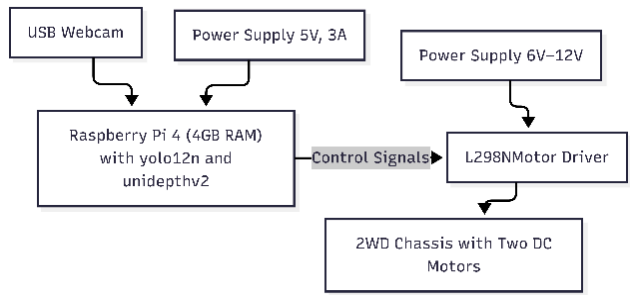

*Figure 2 Hardware Connections*

*1) Hardware Design*
- Compute Unit: Raspberry Pi 4 Model B (4GB RAM), running 64-bit Raspberry Pi OS Bookworm.
- Mobility Platform: A dual-motor rover chassis with BO DC motors controlled via an L298N H-bridge driver powered at 4.8V with 4 AA batteries.
- Vision Sensor: A Logitech C270 USB webcam (720p) was used instead of a stereo setup due to instability and recalibration issues observed during initial field testing.
- Communication: Headless control was enabled via Bluetooth PAN (Personal Area Network), allowing a remote device to connect using VNC for real-time GUI access.

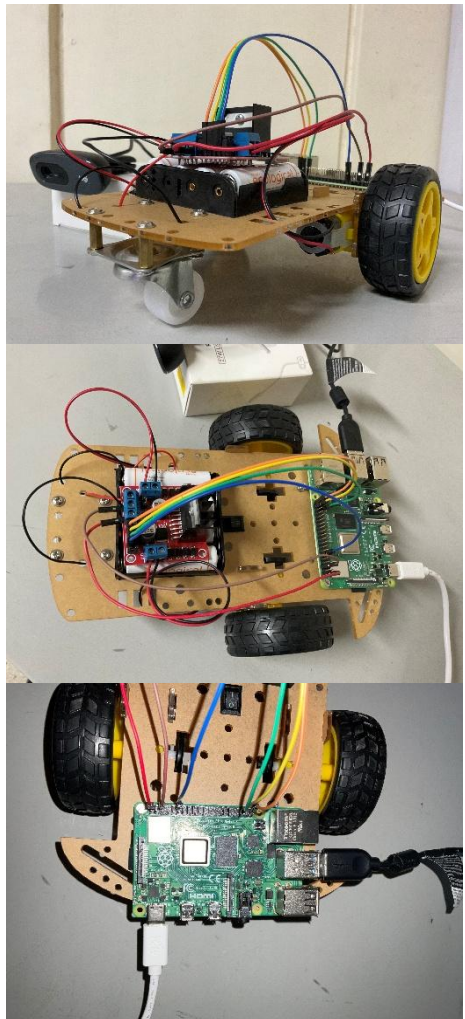

*Figure 3 Images of Rover Prototype*

2) Software Stack

The rover software was developed in Python 3.11 with a modular architecture:

- Depth Estimation: UniDepthV2 (ONNX) [13] was used for monocular metric depth prediction up to 5 meters. Inference runs asynchronously due to the ~7s/frame latency.
- Object Detection: YOLO12n, implemented using Ultralytics [14], achieved ~10 FPS with a confidence threshold of 0.25 and NMS threshold of 0.2.
- Motor Control: GPIO-based controls mapped keyboard inputs (i.e., WASD keys) to motor commands via the L298N driver. PWM is unabled for constant velocity.
- User Interface: A Tkinter-based GUI provided manual override, live video feed, detection overlays, and depth visualization. Like the simulation, the GUI enables to load predefined path that the rover can execute in once. The headless Raspberry Pi and VNC access was enabled using blueman, dnsmasq and RealVNC [15], [16], [17].

Four key evaluation metrics were defined:

- Depth Accuracy: Mean Absolute Error (MAE) against ground-truth distances (0.15 m to 2.0 m) measured using a physical scale.
- Processing Performance: Measured frame rates, inference latency, and system resource utilization (CPU load, memory).
- Navigation Effectiveness: Measured obstacle course completion rate, time taken, and path deviation from intended route.

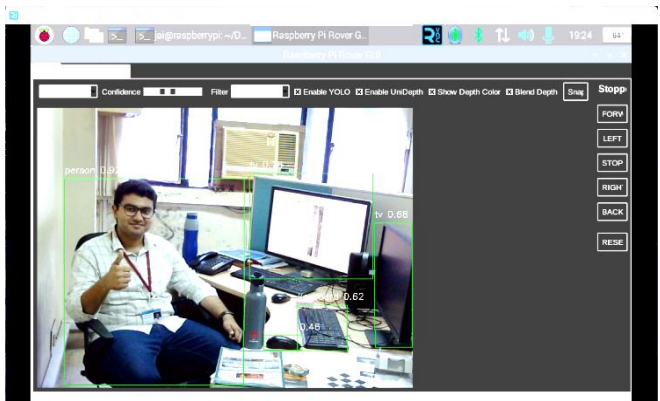

*Figure 4 Real time object detection*

To evaluate the system across varying conditions, experiments were conducted in an indoor laboratory with stable lighting and fixed obstacles, A representative output from the system, shown in Fig. 3, captures real-time object detection and depth estimation using YOLO12n and UniDepthV2 running on Raspberry Pi 4B.

Each scenario was executed multiple times to ensure statistical relevance. During each run, the system logged timestamped RGB images, depth maps, and object detection outputs, along with telemetry data including CPU temperature and memory usage. Navigation trajectories and obstacle interaction events were also recorded for later analysis.

E. System Coordination and Validation

To optimize both safety and performance under limited computational resources, depth estimation ran asynchronously at 0.1 Hz while YOLO12n maintained real-time detection at 10 Hz. This architecture ensured timely obstacle detection while accepting delayed depth feedback for longer-range analysis. Validation included cross-checking depth values, calculating detection precision and recall, and confirming successful autonomous traversal through progressively complex environments.

IV. RESULTS AND DISCUSSION

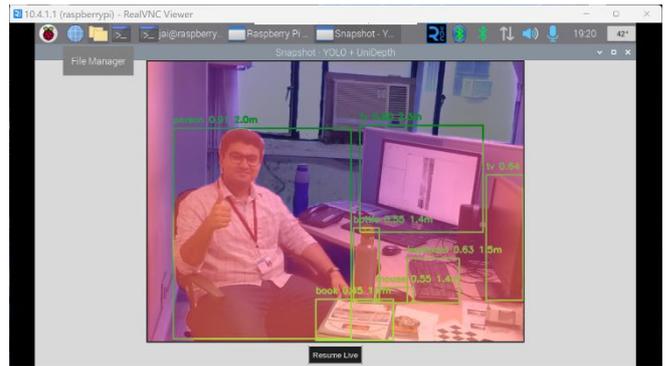

*Figure 5 Objects detected with Metric Depth*

Fig. 5. Output snapshot from the rover's onboard GUI showing simultaneous object detection and monocular depth estimation. YOLO12n detects multiple objects (e.g., person at 2.0 m, bottle at 1.4 m, keyboard at 1.5 m, monitor at 2.3 m) with confidence scores, while UniDepthV2 overlays a color-coded depth map. Estimated distances (in meters) are annotated for each detection. Inference was performed on a Raspberry Pi 4B via RealVNC, with system telemetry (temperature: 42 °C, time: 19:20) shown in the top-right corner. A representative example of the output is shown in Fig. 5, illustrating real-time object detection and monocular depth overlays during live inference. The system achieved practical performance on edge hardware, with the YOLO12n model reaching real-time operation at 10 FPS using the NCNN backend, compared to 1 FPS with the ONNX variant. For depth estimation, the original UniDepthV2 model required over 60 seconds per frame, making it unsuitable for real-time use. In contrast, the optimized ONNX version reduced inference time to approximately 7 seconds per frame with comparable accuracy, allowing deployment in a background thread alongside real-time object detection. With optimizations such as input resolution adjustment and selective quantization, the Raspberry Pi 4B sustained thermal stability, maintaining operating temperatures between 40°C and 65°C, as visible in the corners of Fig. 4 and Fig. 5. These results demonstrate that although continuous real-time depth estimation remains infeasible on low-power devices, a hybrid approach—pairing high-speed detection with slower depth inference—enables effective and safe autonomous navigation. The study also underscores deployment-critical factors such as thermal control, inference scheduling, and accuracy-performance trade-offs that are frequently overlooked in simulation-only evaluations.

| | |
|---|---|
| YOLO12n (NCNN) | 10 FPS |
| YOLO12n (ONNX) | 1 FPS |
| UniDepthV2 (PyTorch) | >60 sec/frame |
| UniDepthV2 (ONNX) | ~7 sec/frame |
| CPU Temperature Range | 40°C–65°C |

*Table 1 summary table for results*

## V. Conclusion and Future Work

This study explored the practical implementation of depth-aware navigation for low-cost autonomous rovers, comparing idealized stereo-based simulation with real-world deployment using monocular vision and edge AI. While stereo vision yielded accurate depth in simulation, real-world challenges such as calibration fragility, hardware noise, and environmental variability highlighted its limitations. By integrating YOLO12n for real-time object detection and an optimized UniDepthV2 ONNX model for asynchronous metric depth inference, we demonstrated a functional hybrid navigation system on a Raspberry Pi 4B. System-level optimizations, including model quantization and thermal management, were essential to maintaining stable operation under resource constraints.

We also plan to integrate sensor fusion techniques—such as combining monocular vision with IMU or stereo fallback—to enhance perception robustness. Another key direction involves implementing a digital twin framework to enable real-time synchronization between simulated and physical rovers, improving development feedback loops and enabling remote mission rehearsal. Field trials in lunar analog terrains and with dynamic obstacles will further evaluate the system's readiness for real-world exploratory scenarios.

## Acknowledgment

The authors extend their sincere gratitude to the team members of Planetary and Space Data Processing Division (PSPD) for their invaluable vision, continuous support, and insightful guidance throughout this work.